\documentclass[runningheads]{llncs}
\usepackage{graphicx}

\usepackage{tikz}
\usepackage{comment}
\usepackage{amsmath,amssymb} 
\usepackage{color}

\usepackage[accsupp]{axessibility}  

\usepackage{tablefootnote}
\usepackage{threeparttable}
\usepackage[T1]{fontenc}
\usepackage{fancyhdr}
\usepackage{graphicx, setspace}
\usepackage{subcaption}
\usepackage{sectsty}
\usepackage{setspace}
\usepackage{url}
\usepackage{enumitem}
\usepackage[normalem]{ulem}
\usepackage{colortbl}
\usepackage{multirow}
\usepackage{booktabs}
\usepackage{bm}
\usepackage{cite}

\begin{document}
\pagestyle{headings}
\mainmatter
\def\ECCVSubNumber{10}  

\title{Artifact-based Domain Generalization\\of Skin Lesion Models} %

%
\author{Alceu Bissoto\orcidID{0000-0003-2293-6160}\inst{1,4} \and
Catarina Barata\orcidID{0000-0002-2852-7723}\inst{2} \and
Eduardo Valle\orcidID{0000-0001-5396-9868}\inst{3,4} \and Sandra Avila\orcidID{0000-0001-9068-938X}\inst{1,4}}
\authorrunning{Bissoto et al.}
%
\institute{Institute of Computing, University of Campinas, Brazil \\ \email{\{alceubissoto, sandra\}@ic.unicamp.br}
\and Institute for Systems and Robotics, Instituto Superior Técnico, Portugal \email{ana.c.fidalgo.barata@tecnico.ulisboa.pt}
\and School of Electrical and Computing Engineering, University of Campinas, Brazil \email{dovalle@dca.fee.unicamp.br}
\and Recod.ai Lab, University of Campinas, Brazil}
\maketitle

\begin{abstract}
Deep Learning failure cases are abundant, particularly in the medical area.
Recent studies in out-of-distribution generalization have advanced considerably on well-controlled synthetic datasets, but they do not represent medical imaging contexts.
We propose a pipeline that relies on artifacts annotation to enable generalization evaluation and debiasing for the challenging skin lesion analysis context. 
First, we partition the data into levels of increasingly higher biased training and test sets for better generalization assessment. Then, we create environments based on skin lesion artifacts to enable domain generalization methods. Finally, after robust training, we perform a test-time debiasing procedure, reducing spurious features in inference images.
Our experiments show our pipeline improves performance metrics in biased cases, and avoids artifacts when using explanation methods. 
Still, when evaluating such models in out-of-distribution data, they did not prefer clinically-meaningful features. Instead, performance only improved in test sets that present similar artifacts from training, suggesting models learned to ignore the known set of artifacts. Our results raise a concern that debiasing models towards a single aspect may not be enough for fair skin lesion analysis.

\end{abstract}

\section{Introduction}
Despite Deep Learning's superhuman performance on many tasks, models still struggle to generalize, stalling the adoption of AI for critical decisions such as medical diagnosis. 

Skin lesion analysis is no exception. Recent works exposed concerning  model behaviors, such as achieving high performances with the lesions fully occluded on the image \cite{bissoto2019constructing}, or exploiting the presence of artifacts (e.g., rulers positioned by dermatologists to measure lesions) to shortcut learning \cite{bissoto2020debiasing}. Moreover, current models fail to cope with underrepresented populations such as Black, Hispanic, and Asian people. Those shortcomings prevent automated skin analysis solutions from wider adoption and from realizing their potential public health benefits.

Domain Generalization (DG), which in computer vision studies how models fall prey to spurious correlations, is yet to be adequately adopted by the medical image analysis literature, partly because medical data often lack the \textit{labeled environments} which are a critical input to most DG techniques. Within a corpus of data, Environments are groups or domains that share a common characteristic (e.g., predominant image color, image capturing device, demographic similarities). In DG research, datasets are often synthetic, creating environments on demand, or multi-sourced, with an environment for each source. Medical data, however, pose special challenges due to their complexity and multi-faceted nature, presenting multiple ways of grouping data, or latent environments whose full annotation is next to impossible. We are interested in adapting DG techniques to benefit those complex and rich tasks, considering those challenges.

We start by allowing the assessment of generalization performance, even when out-of-distribution data are unavailable, using a tunable version of “trap sets” \cite{bissoto2020debiasing}. Next, we infer existing, latent environments from available data, enabling the adoption of robust learning methods developed in the DG literature. Finally, after model training, we select robust features during test-time, censoring irrelevant information. Our extensive experiments show that it is possible to obtain models that are resilient to training with highly biased data. Code to reproduce our experiments is available at \url{https://github.com/alceubissoto/artifact-generalization-skin}. 

Our main contributions are:
\begin{itemize}
    \item We propose a method to adapt existing annotations into environments, successfully increasing the robustness of skin lesion analysis models;
    \item We propose a test-time procedure that consistently improves biased models' performance;
    \item We show that model debiasing is insufficient to increase out-of-distribution performance. Better characterization of out-of-distribution spurious sources is necessary to train more robust models.
\end{itemize}

\section{Background}

Domain Generalization (DG) and Domain Adaptation (DA) aim to study and mitigate distribution shifts between training and test data for a known (in the case of DA) or an unknown (in the case of DG) distribution test data distribution. Here we focus on DG techniques since the test distribution is almost always unknown for medical analysis.

A complete review of the extensive literature on DG is outside the scope of this work. We point the reader to two recent surveys of the area \cite{zhou2021domain,shen2021towards}. In this section, we will briefly review the two techniques directly used in this work.

DG techniques are contrasted with the classical \textbf{Empirical Risk Minimization \cite{vapnik1992principles} (ERM)} learning criterion, which assumes that the samples are independent and identically distributed (i.i.d.) and that train and test sets are sampled from the same distribution. For the sake of completeness, the ERM minimization goal is defined as 
$R_{\mathrm{ERM}}(\theta)=\frac{1}{n} \sum_{i=1}^{n} \ell\left(x_{i}, y_{i} ; \theta\right)$,
where $\ell$ is the classification loss, $\theta$ is the model's parameters, and $n$ is the number of samples~$(x,y)$.
DG techniques deal with train–test distribution shifts. We present two of them below. 

\textbf{Distributional Robust Optimization (DRO)} \cite{hu2018does, sagawa2019distributionally} methods minimize the maximum risk for all groups (while ERM minimizes the global average risk). That way, the model focuses on high-risk groups, which usually comprise those with correlations underrepresented in the dataset. The risk is calculated as:
\begin{equation}
R_{\mathrm{DRO}}(\theta):=\max _{e \in \mathcal{E}_{\mathrm{tr}}} \hat{\mathbb{E}}_{P^{e}}[\ell(x, y;\theta)] ,
\end{equation}
\noindent where we evaluate the expectation separately for each environment distribution $P^{e}$, and the data is separated into environments $e$, sampled from the set of all environments available for training $\mathcal{E}_{\mathrm{tr}}$.

DRO can prevent models from exploiting spurious correlations, for example, if the risk is low for a biased group and high for an unbiased group. In that case, success depends on groups being separated by bias. DRO can also raise the importance of small groups (e.g., rare animal subspecies, rare pathological conditions), which would be obliterated by averaging. DRO techniques require explicitly labeled environments, and one of our main contributions is evaluating one of them (GroupDRO) on inferred environments.

\textbf{Representation Self-Challenging (RSC) \cite{huang2020self}} is a three-step robust deep learning training method. At each training iteration, RSC sets to zero the most predictive part of the model representation, according to the gradients.
More specifically, the model representations with the highest gradients will be set to zero before the model update.
Such feature selection causes less dominant features in the training set to be learned by the model, potentially discarding easy-to-learn spurious correlations and thus preventing the so-called \textit{shortcut learning} \cite{shah2020pitfalls}. 
We use RSC as a strong baseline for comparing with our proposed pipeline, since this technique does not require environment labels, being adaptable to any classification problem. In a recent benchmark \cite{ye2022ood}, RSC appears as one of the few effective methods, including for the PatchCamelyon histopathology dataset~\cite{bandi2018detection, koh2020wilds}.

\section{Methodology}

\begin{figure}[th]
    \centering
    \includegraphics[trim={0.5cm 0.25cm 0.5cm 0.5cm},clip,width=\linewidth]{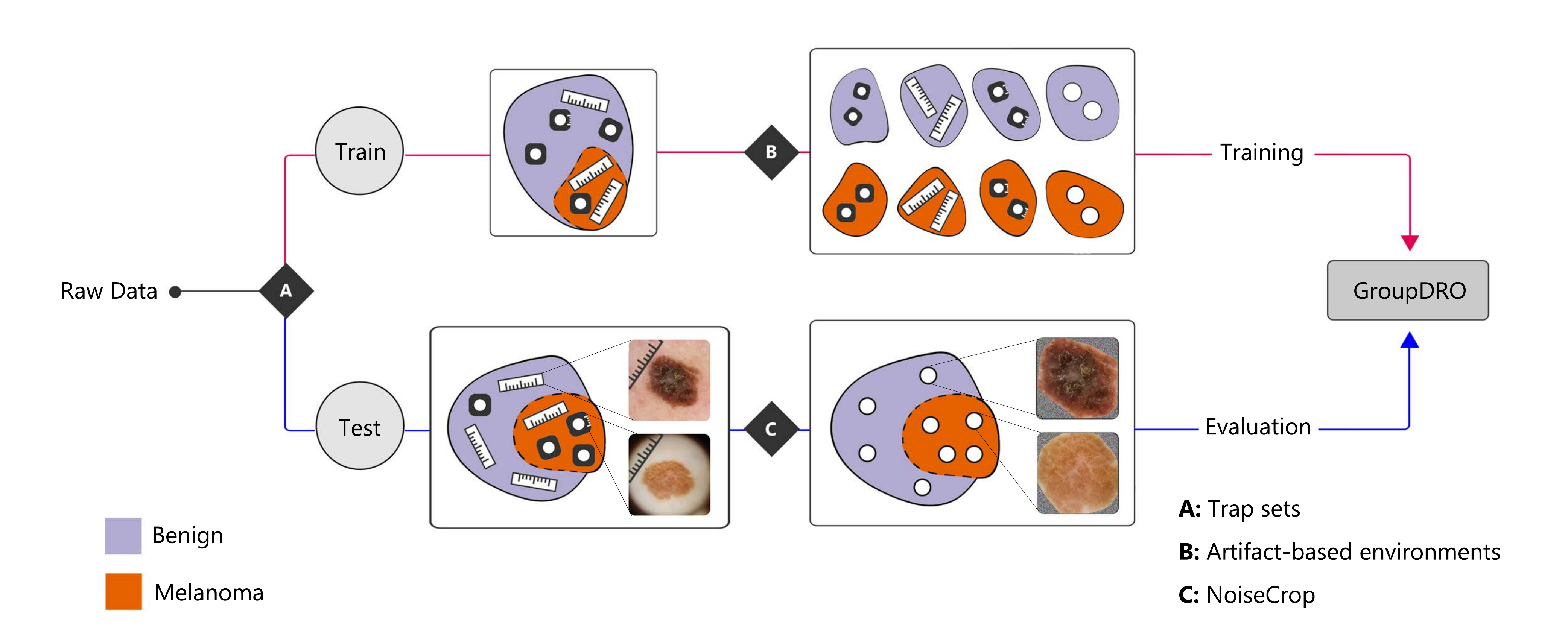}
    \caption{Our proposed pipeline for debiasing. 
Images with artifacts are represented by the drawings of rulers, dark corners, and the combination of both. White circles represent samples where artifacts are absent. Our first step (A) partitions data into challenging train and test sets, called Trap sets. The training set is divided into environments (B). Each environment groups samples containing the same set of artifacts. These environments are used to train a robust learning algorithm, such as GroupDRO. In the last step (C), we select features of our trap-test set, censuring the background which may provide spurious correlations. } 
    \label{fig:intropipeline}
\end{figure}

The main objective is to learn more robust skin lesion representations using deep learning for skin lesion analysis, considering the binary problem of melanoma \textit{vs.} benign. To achieve this, we present a pipeline (Fig.~\ref{fig:intropipeline}) that proposes 1)~partitioning data into train/test trap sets that simulate a highly biased scenario; 2)~crafting and exploiting training data partitions (environments) to learn robust representations through GroupDRO \cite{sagawa2019distributionally}; 3) selecting task-relevant features for inference, avoiding spurious ones.

\subsection{Trap Sets} \label{sec:trap}

Since spurious correlations inflate metrics, DG methods require carefully crafted protocols to measure generalization. Often, datasets introduce correlations with the class labels (using an extraneous feature, such as color in ColorMNIST~\cite{arjovsky2019invariant}), which purposefully differ between the training and the test split.

Here, we follow the ``trap set'' procedure~\cite{bissoto2020debiasing} to craft training and test with \textbf{amplified} correlations between artifacts and class labels (malignant \textit{vs.} benign), which appear in \textbf{opposite} directions in the dataset splits. We adapt the trap set protocol, introducing a tunable level of bias, from 0 (randomly selected sets) to 1 (highly biased). This level controls, for each sample in the split, the probability of selecting it at random \textit{versus} following the trap set procedure. Table~\ref{tab:corrtraps} illustrates the correlations between artifact and class labels on the splits for the bias levels used in this work. 
We think our trap sets can expand generalization measurements to be used outside of specialized literature, reaching problems that urgently need out-of-distribution performance assessment.

\begin{table*}[htb]
\scriptsize
\begin{center}
    \caption{Spearman correlations between diagnostic and each of the $7$ considered artifacts to build the trap sets. As the factor increase, so does the correlations and differences between train and test.}
    \begin{tabular}{ccccccccc}
        factor & set & dark corner & hair & gel border & gel bubble & ruler & ink & patches  \\
        \toprule
        $0$ &train &\cellcolor[HTML]{d8efe4}0.119 &\cellcolor[HTML]{f9e3e1}-0.104 &\cellcolor[HTML]{feffff}0.003 &\cellcolor[HTML]{edf8f3}0.055 &\cellcolor[HTML]{d0ecde}0.142 &\cellcolor[HTML]{f8fcfa}0.023 &\cellcolor[HTML]{f8dad8}-0.138 \\
~ &test &\cellcolor[HTML]{d2ede0}0.135 &\cellcolor[HTML]{f9e1df}-0.112 &\cellcolor[HTML]{f8fcfa}0.023 &\cellcolor[HTML]{f0f9f5}0.047 &\cellcolor[HTML]{c9e9da}0.162 &\cellcolor[HTML]{f5fbf8}0.030 &\cellcolor[HTML]{f7d7d5}-0.149 \\
\midrule
$0.5$ &train &\cellcolor[HTML]{b1e0c9}0.233 &\cellcolor[HTML]{f5cecb}-0.185 &\cellcolor[HTML]{e4f4ec}0.083 &\cellcolor[HTML]{eef8f3}0.052 &\cellcolor[HTML]{addec6}0.246 &\cellcolor[HTML]{eff9f4}0.048 &\cellcolor[HTML]{f9e2e0}-0.110 \\
~ &test &\cellcolor[HTML]{f8ddda}-0.129 &\cellcolor[HTML]{e4f4ec}0.083 &\cellcolor[HTML]{f7d6d3}-0.156 &\cellcolor[HTML]{f3faf7}0.038 &\cellcolor[HTML]{fbebea}-0.074 &\cellcolor[HTML]{fdf8f8}-0.025 &\cellcolor[HTML]{f4c6c2}-0.217 \\
\midrule
$1.0$ &train &\cellcolor[HTML]{87cfab}0.36 &\cellcolor[HTML]{f0b5b0}-0.282 &\cellcolor[HTML]{c4e7d6}0.178 &\cellcolor[HTML]{edf8f2}0.056 &\cellcolor[HTML]{89d0ad}0.352 &\cellcolor[HTML]{dff2e9}0.096 &\cellcolor[HTML]{fbeeed}-0.062 \\
~ &test &\cellcolor[HTML]{e98c84}-0.438 &\cellcolor[HTML]{9cd7ba}0.296 &\cellcolor[HTML]{eda39d}-0.35 &\cellcolor[HTML]{eff9f4}0.049 &\cellcolor[HTML]{eea7a1}-0.335 &\cellcolor[HTML]{f9e1df}-0.113 &\cellcolor[HTML]{efaba5}-0.319 \\
        \bottomrule
    \label{tab:corrtraps}     
    \end{tabular}
    \end{center}
\end{table*}

\subsection{Artifact-based Environments} \label{sec33:envs}

In  DG, environments divide data according to spurious characteristics. For example, ColorMNIST \cite{arjovsky2019invariant} is divided into two environments: one that correlates colors with values (one color for each digit), and another with colors chosen randomly.
Some environments correspond to data sources, as in PatchCamelyon \cite{bandi2018detection, koh2020wilds}, where each environment comprises data collected at the same hospital.

Arjovsky et al. \cite{arjovsky2019invariant} mention that environments act to ``reduce degrees of freedom in the space of invariant solutions''. Thus, more environments help discard spurious features during training \cite{rosenfeld2020risks}. The plethora of concepts available enables multiple ways of dividing the dataset into environments, some of which will be more successful than others at achieving robust representations.

Many annotated concepts could be used for environment generation for skin lesion datasets. Recently, Daneshjou et al.~\cite{daneshjou2021disparities} released a clinical skin lesion dataset presenting per-image specialist annotated information on Fitzpatrick skin types. Other metadata such as anatomical location, patient sex, and age are available for some datasets, such as the ISIC2019 \cite{combalia2022validation}.
In this work, we use the presence of artifacts in the image capture process to create the environment. The presence of those artifacts gives models the opportunity to exploit spurious correlations in order to shortcut learning \cite{bissoto2020debiasing, winkler2019association, combalia2022validation}. We aim to prevent that, creating more robust models.

The $7$ artifact types (see Table \ref{tab:corrtraps}) may co-occur in lesions, with $2^7=128$ combinations. Adding the binary class label, that gives $256$ potential environments (e.g., benign with no artifacts, benign with dark-corners, malignant with dark-corners and rulers, etc.), although some of those may contain very few (or zero) images. We use non-empty environments to train a robust learning algorithm. 

Our risk minimization of choice is Group Distributionally Robust Optimization (GroupDRO)~\cite{sagawa2019distributionally}, a variation of DRO that includes more aggressive regularization, in the form of a hyperparameter to encourage fitting smaller groups, higher $\ell_2$ regularization, and early stopping. It is a good fit due to our setting  with many few-samples environments.

\subsection{NoiseCrop: Test-time Feature Selection} \label{sec36:feat}

The last step in the pipeline is selecting robust features for inference. Recent work \cite{borji2021contemplating} shows that test-time feature selection yields considerable gains in performance, even when spurious correlations are learned.

In this step, we censor the input images' information to prevent models from using spurious features. We employ segmentation masks to separate foreground lesions, which host robust features, from background skin areas, which concentrate spurious information (e.g., skin tones, patches, and image artifacts).

\begin{figure}[htb]
    \centering
    \begin{subfigure}[b]{0.25\linewidth}
    \includegraphics[width=\linewidth,height=\linewidth]{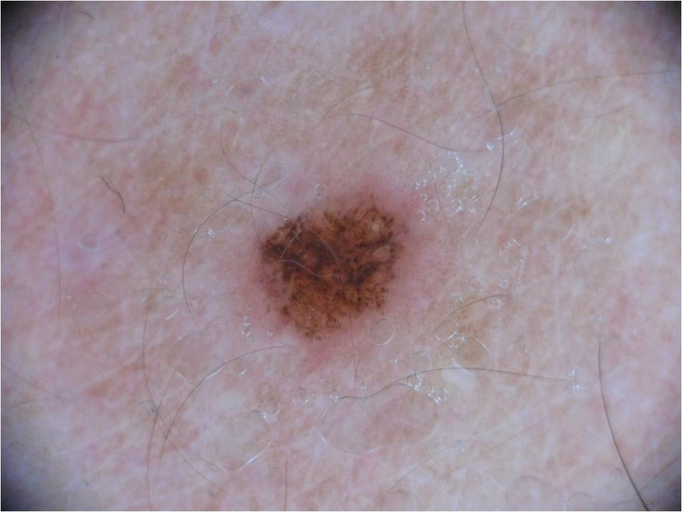}
    \caption{Original}
    \end{subfigure}
    \begin{subfigure}[b]{0.25\linewidth}
    \includegraphics[width=\linewidth]{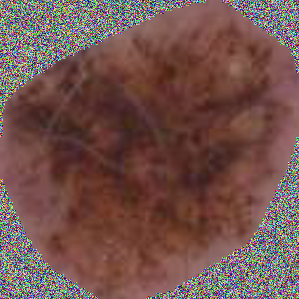}
    \caption{NoiseCrop}
    \end{subfigure}
    \caption{Comparison between Original and NoiseCrop images. In NoiseCrop, we remove the background information,  replace it with a uniform noise, and resize the lesion to occupy the whole image.}
    \label{fig:normalizedimages}
\end{figure}

We employ the ground-truth segmentation masks when available, and infer the segmentation (with a Deep Learning model~\cite{chen2018encoder}) when they are not.
Since we post-process all masks through a convex hull operation, masks do not need to be pixel-perfect, instead they must roughly cover the whole lesion. 
To minimize the effect of the background pixels on the models, we replace them with a noisy background sampled uniformly from 0 to 255 in each RGB channel.
We also eliminate \textit{lesion size} information since the lack of scale guidelines for image capture makes size an unreliable feature subjected to spurious correlations.
The convex hull of the segmentation mask is used to crop and re-scale the image such that lesion occupies the largest possible area while keeping the aspect ratio.
We call those censoring procedures \textbf{NoiseCrop} (Fig.~\ref{fig:normalizedimages}). Again, we stress, this censoring is applied only to test images.

\section{Results}

\subsection{Data}

We employ several high-quality datasets in this study (Table \ref{tab:datasets}). The class labels are selected and grouped such that the task is always a binary classification of melanoma \textit{vs.} benign (other, except for carcinomas). We removed from all analysis samples labeled basal cell carcinoma or squamous cell carcinoma. In the out-of-distribution test sets, we kept only samples labeled melanoma, nevus, and benign/seborrheic keratosis.

\begingroup
\setlength{\tabcolsep}{2pt} 
\renewcommand{\arraystretch}{1.4} 
\begin{table*}[t]
\begin{center}
\caption{Datasets used in our work}
\scriptsize
\begin{tabular}{p{2.7cm}>{\raggedleft}p{1.7cm}>{\raggedright}p{4.1cm}>{\raggedright}p{1.2cm}>{\raggedright\arraybackslash}p{1.5cm}}
\toprule

Dataset & \# Samples & Classes  & Set & Type \\
\midrule
ISIC2019 (train)~\cite{codella2019skin} & 12,360    & melanoma \textit{vs.} nevus, actinic keratosis, benign keratosis, dermatofibroma, vascular lesion & training                & dermoscopic \\
ISIC2019 (val)~\cite{codella2019skin} & 2,060    & as above & validation                & dermoscopic \\
ISIC2019 (test)~\cite{codella2019skin} & 6,182    & as above & test                & dermoscopic \\
PH2~\cite{mendoncca2015ph2}                 & 200        & melanoma \textit{vs.} nevus, benign keratosis & test & dermoscopic  \\
Derm7pt-Dermoscopic~\cite{Kawahara2018-7pt}             & 872      & as above  & test & dermoscopic \\
Derm7pt-Clinical~\cite{Kawahara2018-7pt}             & 839      & as above  & test & clinical \\
PAD-UFES-20~\cite{pacheco2020pad}         & 531     & as above                                  & test & clinical              \\
\bottomrule
\end{tabular}

\label{tab:datasets}
\end{center}
\end{table*}
\endgroup

The artifact annotations \cite{bissoto2020debiasing} comprise $7$ types: \textit{dark corners} (\textit{vignetting}), \textit{hair, gel borders, gel bubbles, rulers, ink markings/staining}, and \textit{patches} applied to the patient skin. Ground-truth labels for those are available for the ISIC2018~\cite{isic2018data} and Derm7pt~\cite{Kawahara2018-7pt}. For the larger ISIC2019, we infer those labels using independent binary per-artifact classifiers fine-tuned on the ISIC2018 annotations\footnote{Each model is an ImageNet-pretrained Inceptionv4~\cite{szegedy2016inceptionv4} fine tuned with stochastic gradient descent, with momentum 0.9, weight decay $10^{-3}$, and learning rate $10^{-3}$, reduced to $10^{-4}$ after epoch 25. Batch size is 32, with reshuffling before each epoch. Data augmented with random crops, rotations, flips, and color transformations.}.

\subsection{Model Selection and Implementation Details} 
Hyperparameter selection is crucial for DG.
Following GroupDRO \cite{sagawa2019distributionally} protocol, we first performed a grid-search over learning rate (values 0.00001, 0.0001, 0.001), and weight-decay (0.001, 0.01, 0.1, 1.0), for 2 runs, on a validation set randomly split from the training set. Although GroupDRO suggests an unbiased (equal presence of all artifacts) validation set, we found such constraint unrealistic, since a perfectly unbiased data distribution is impossible to predict at training time. We follow the same hyperparameter search procedures for all techniques, including the baselines.
Given the best combination on the validation set, we searched for GroupDRO's generalization adjustment argument among the values [0..5]. Sagawa et al. \cite{sagawa2019distributionally} added that hyperparameter to encourage fitting smaller groups. We provide, to illustrate an upper-bound of GroupDRO's performance, an \textbf{oracle} version whose hyperparameters were selected with privileged information from test time.

All models employ a ResNet-50 \cite{he2016deep} backbone, fine-tuned for up to 100 epochs with SGD with momentum and patience of 22 epochs. Conventional data augmentation (shifts, rotations, color) is used on training and testing, with 50 replicas for the latter. On all plots, lines refer to the average of 10 runs, with shaded areas showing the standard error. Each run has a different training/validation partition and random seed.

\subsection{Debiasing of Skin Lesion Models}

The trap set protocol partitions train and test in an intentional challenging way that is catastrophic for naive models. Models that exploit spurious correlations in the train “fall in the trap” resulting in very low performance (Table~\ref{tab:main_results}).

ERM achieves a ROC AUC of only $0.58$, showing that trap sets successfully creates challenging biased train and test sets \cite{bissoto2020debiasing}.
Our pipeline consider GroupDRO enabled by our artifact-based environments, followed by the application of NoiseCrop in test images. Debiased methods should produce solutions that are more invariant to the training bias, varying less from low to high bias scenarios. 

\begin{table}[bht]
    \begin{center}
    \caption{Results for different pipelines on a strong trap test (training bias = 1). Our results considerably surpass the state of the art in that scenario.\\
    \footnotesize{\dag Reported from the original, using a ResNet-152 model on the ISIC2018 dataset.}}
    \footnotesize
    \begin{tabular}{lc}
        \toprule
         Method & ROC AUC \\
         \midrule
         ERM~\cite{vapnik1992principles} & $0.58$ \\
         RSC~\cite{huang2020self} & $0.59$\\
         Bissoto et al.~\cite{bissoto2020debiasing}$^\dag$ & $0.54$ \\ 
         GroupDRO (Ours) & $0.68$ \\
         Full Pipeline (Ours) & \bm{$0.74$} \\
         \bottomrule
    \end{tabular}
    
    \label{tab:main_results}
    \end{center}
\end{table}

Our solution reaches $0.74$ AUC in the most biased scenario, while the ERM baseline performs not much better than chance --- a difference of $16$ percentage points.
Other robust methods that do not make use of environments (RSC and Bissoto et al. \cite{bissoto2020debiasing}) failed to improve over ERM. To the best of our knowledge, this is the first time debiasing solutions succeed for skin lesion analysis. 

\noindent\textbf{Summary:} 
Our pipeline is an effective strategy for debiasing, surpassing baselines and previous works by 16 percentage points in high-bias scenarios. 

\subsection{Ablation Study}

\begin{figure}[h]
    \centering
    \includegraphics[width=0.9\linewidth]{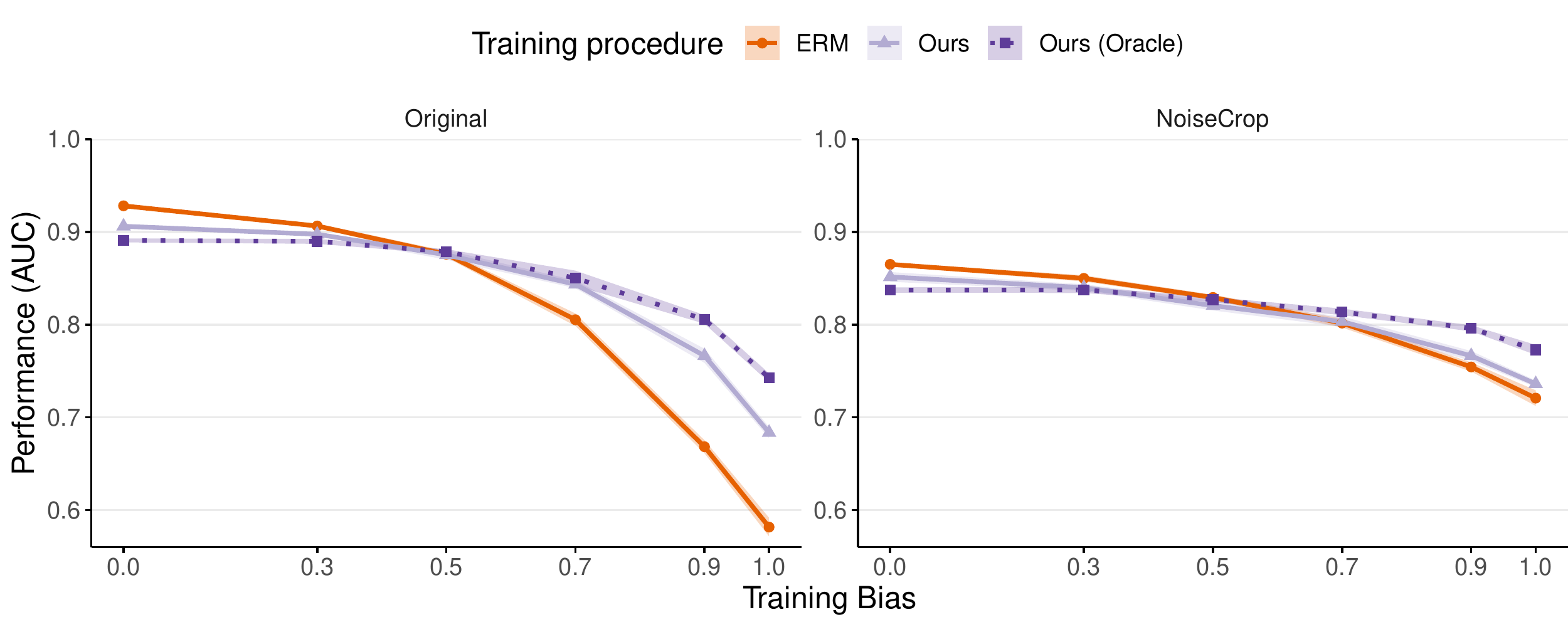}
    \caption{Ablation study of our method. Each line represent a training method. On the left, we perform inference with original test samples, and in the right, we use NoiseCrop for inference. The Oracle serves as an upper-bound, where we ran our pipeline with access to the test distribution for hyperparameter decision. All methods are evaluated at our trap sets with increasing bias.
     Our artifact-based environments enable GroupDRO to improve robustness, and NoiseCrop improved robustness of all methods.}
    \label{fig:trad_groupdro}
\end{figure}

Next, we provide an ablation of our pipeline, individually evaluating the effects of the robust training enabled by our artifact-based environments and NoiseCrop. We consider increasingly high training biases to check the differences of performances in low and high biased scenarios. We show our results in Fig.~\ref{fig:trad_groupdro}: performances on the right inferred over NoiseCrop images, and without it (original images) on the left.\vspace{-0.25cm}  

\paragraph{Artifact-based GroupDRO} 

GroupDRO increases the robustness to artifacts, yielding an improvement of around $10$ percentage points in the AUC metric for high-bias scenarios. In such biased contexts, trap sets punish the model for relying on the artifacts, causing both ERM and RSC to fall under $0.6$ AUC. 
In low-bias scenarios, GroupDRO prevents models from relying on artifacts, causing the performance to drop compared to the baseline. 
When using privileged information to select hyperparameters for GroupDRO, our oracle reached $0.77$ AUC. In the DG literature, deciding hyperparameters is a crucial step, and it is not uncommon to see methods completely fail when hyperparameters are chosen without privileged information over unbiased sets \cite{gulrajani2020search, ahmed2020systematic}. We believe that our fine-grained environments considering each possible combination of artifacts allowed for more robustness to hyperparameter decision.

\vspace{-0.25cm}
\paragraph{Test-time debiasing}

To complete our proposed pipeline (as Fig.~\ref{fig:intropipeline}), we perform feature selection on inference-time. Unlike the direction usually pursued in the literature \cite{sagawa2019distributionally, arjovsky2019invariant}, our debiasing method does not require altering any procedure during training. The idea is to select the features present in the image \textit{during test evaluation}, forcing the network to use the correct correlations learned to make the prediction. 
In Fig.~\ref{fig:trad_groupdro} (right), the scenario drastically changes when the same networks from the left of the figure are tested with NoiseCrop images, especially for the most biased scenario. 
The ERM model, which was slightly better than chance when classifying skin lesions with unchanged images, surpassed $0.72$ AUC when evaluated with NoiseCrop images.
Composing this test procedure alongside robust training methods further improves performance, achieving our best result. 
The reported harm in the performance for less biased scenarios can be illusory since exploiting biases naturally translates to better in-distribution performance but less generalization power.

The steep increase in performance when using NoiseCrop test samples with the baseline model suggests that the network learns correct correlations even when training is heavily contaminated with spurious correlations, contrary to previous belief \cite{pezeshki2020gradient}.
Still, we achieve our highest performance by using the debiasing procedure (through GroupDRO) and the NoiseCrop test. As in our pipeline, training and test-time debiasing are necessary to create more robust models. 
Test-time debiasing appears as a quick effective method to increase robustness at the cost of using domain knowledge of the task. The main challenge is to make test-time debiasing more general, relying less on existing annotations, such as the segmentation masks we use for skin lesion images. 

\noindent\textbf{Summary:} Artifact-based GroupDRO is an effective strategy for debiasing, and masking artifacts (spurious correlations) during test enable correct features to be used for inference. Our ablation suggests that models still learn robust predictive features even when trained on highly-biased data, but are ignored when known spurious correlations appear during test-time.

\subsection{Out-of-distribution Evaluation}

\begin{figure}[h]
    \centering
    \includegraphics[width=1.0\linewidth]{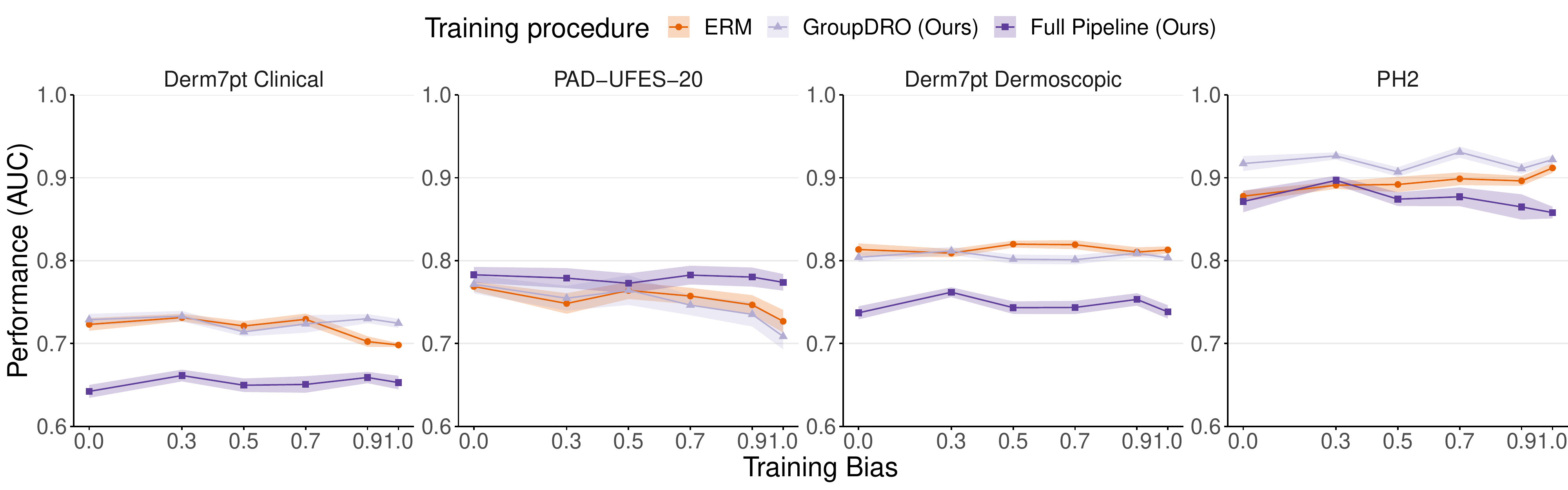}
    \caption{The different lines compare the ERM baseline, our environment-enabled GroupDRO, and our full pipeline. We train the models with increasingly high biased sets (trap train). We evaluate the performance on $4$ out-of-distribution test sets comprising clinical and dermoscopic samples. Unlike the plots using trap test for evaluation, trends here are subtler. The debiasing procedure improves performances on PH2, PAD-UFES-20, and for biased models on Derm7pt-Clinical. On Derm7pt-Dermoscopic, baselines still perform better, despite all the bias in train.}
    \label{fig:ood}
\end{figure}

We have previously shown the increased robustness of skin lesion analysis models when training with our artifact-based environments and NoiseCrop test samples. 
Now, we investigate the effect of the acquired robustness on out-of-distribution sets, which present different artifacts and attributes. Does robustness to the annotated artifacts cause models to rely more on robust features in general?
We show our results in Fig.~\ref{fig:ood}.

The performances on out-of-distribution test sets are more stable than on trap tests across training biases. This is because trap-test contains opposite correlations from training, punishing the model for learning the encouraged spurious correlations. Still, PH2 and PAD-UFES-20 lines show slight negative and positive trends, respectively, indicating the presence and exploitation of biases. 

Our results show very noisy out-of-distribution performance according to the technique used. Our full pipeline present consistent advantage for PAD-UFES-20, while presenting lower performances at all other cases. 
Interestingly, when we skip NoiseCrop, using only GroupDRO for debiasing, we achieve positive results for all training biases in PH2, and for high training biases in Derm7pt Clinical. For Derm7pt-Dermato, the robust training procedure yielded no gains.

\begin{figure}[ht]
  \centering 
  \begin{subfigure}{0.23\linewidth}
  \includegraphics[width=0.48\linewidth, height=0.48\linewidth]{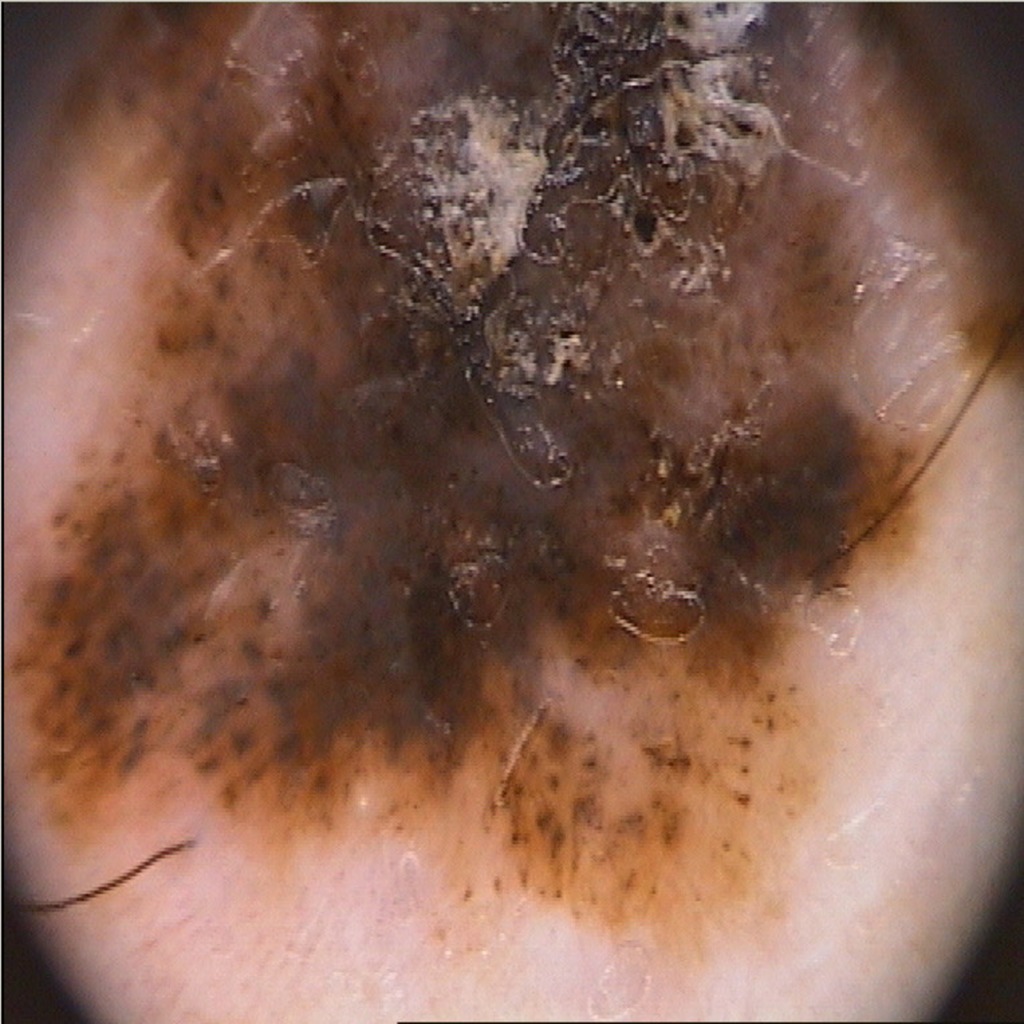}
  \includegraphics[width=0.48\linewidth, height=0.48\linewidth]{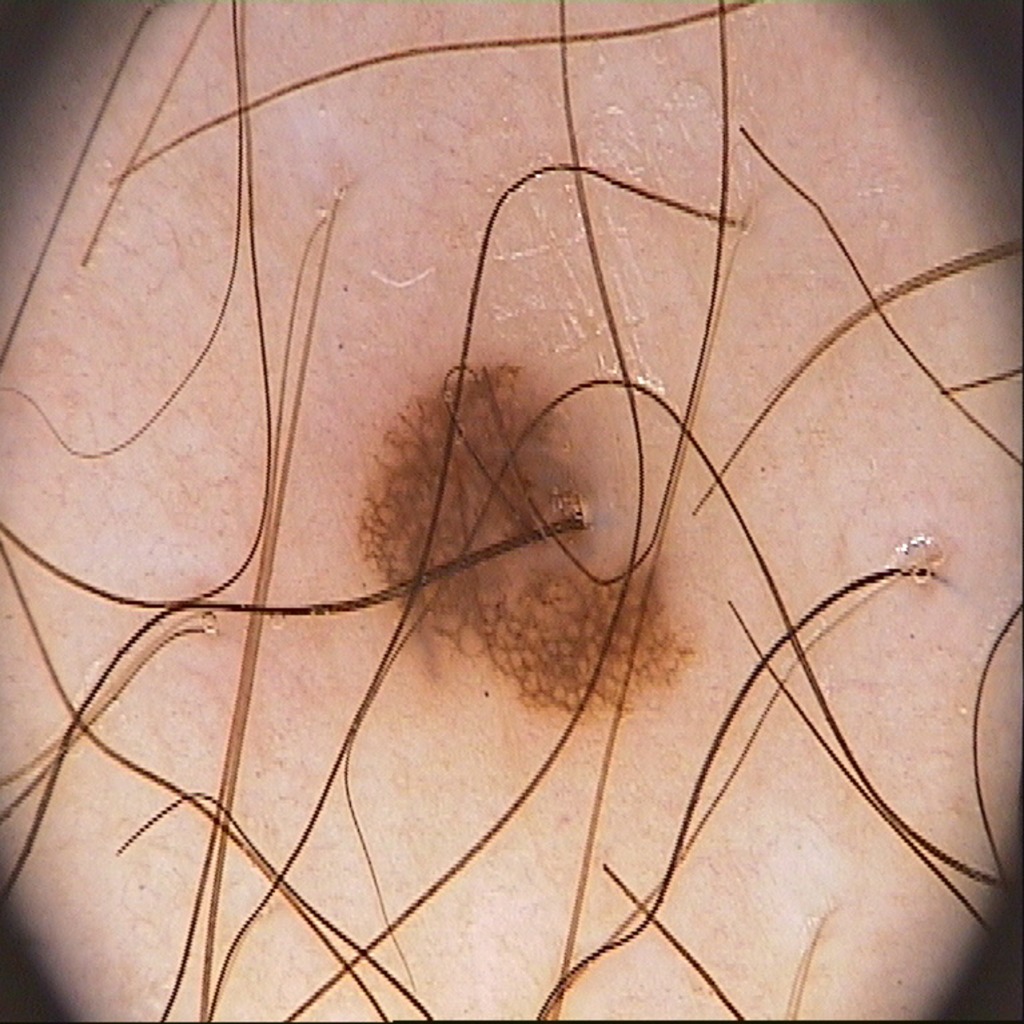}
  \caption{}
  \end{subfigure}\hspace{0.15cm}
  \begin{subfigure}{0.23\linewidth}
  \includegraphics[width=0.48\linewidth, height=0.48\linewidth]{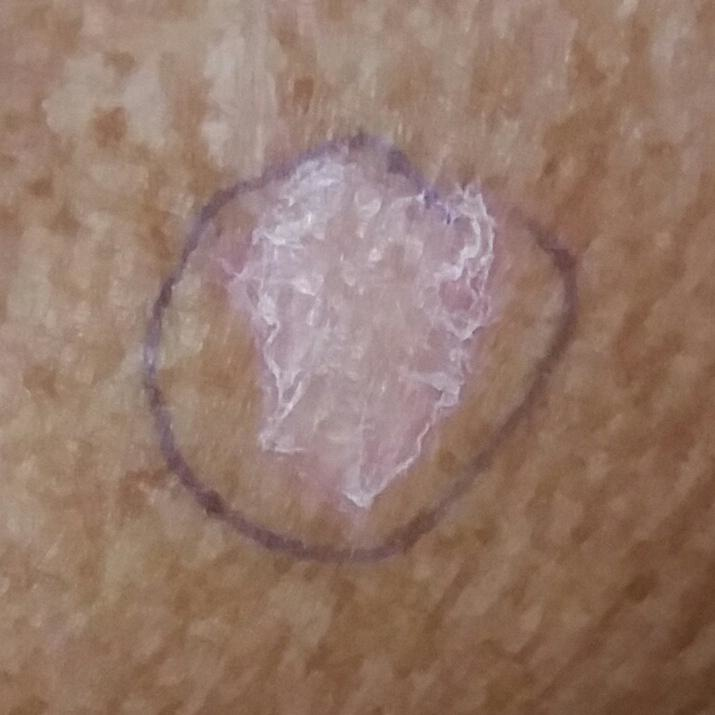}
  \includegraphics[width=0.48\linewidth, height=0.48\linewidth]{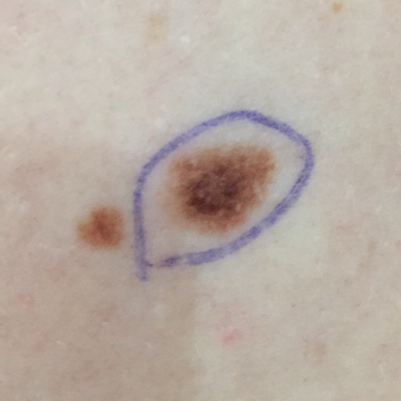}
  \caption{}
  \end{subfigure}\hspace{0.15cm}
  \begin{subfigure}{0.23\linewidth}
  \includegraphics[width=0.48\linewidth, height=0.48\linewidth]{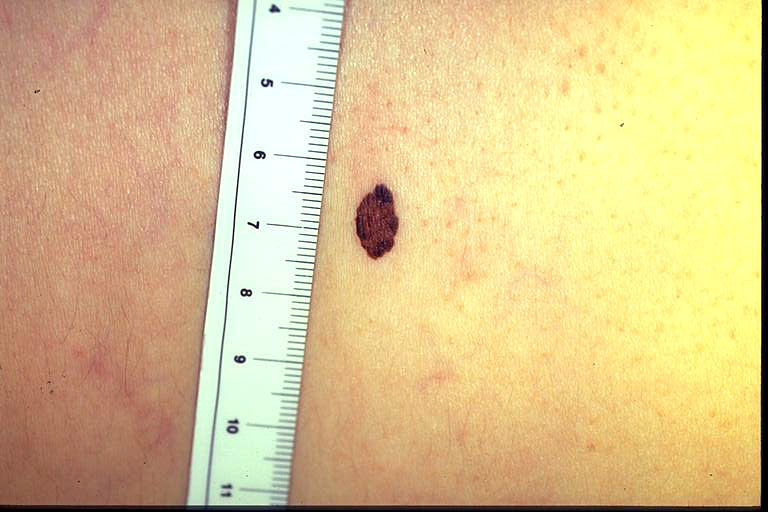}
  \includegraphics[width=0.48\linewidth, height=0.48\linewidth]{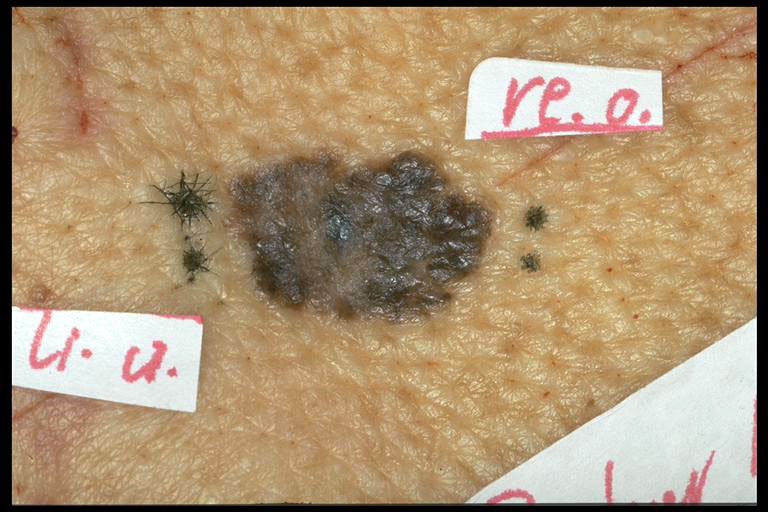}
  \caption{}
  \end{subfigure}\hspace{0.15cm}
  \begin{subfigure}{0.23\linewidth}
  \includegraphics[width=0.48\linewidth, height=0.48\linewidth]{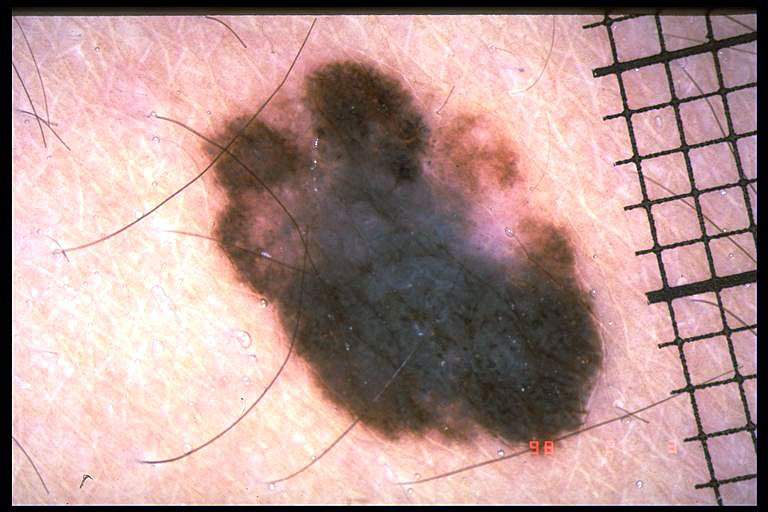}
  \includegraphics[width=0.48\linewidth, height=0.48\linewidth]{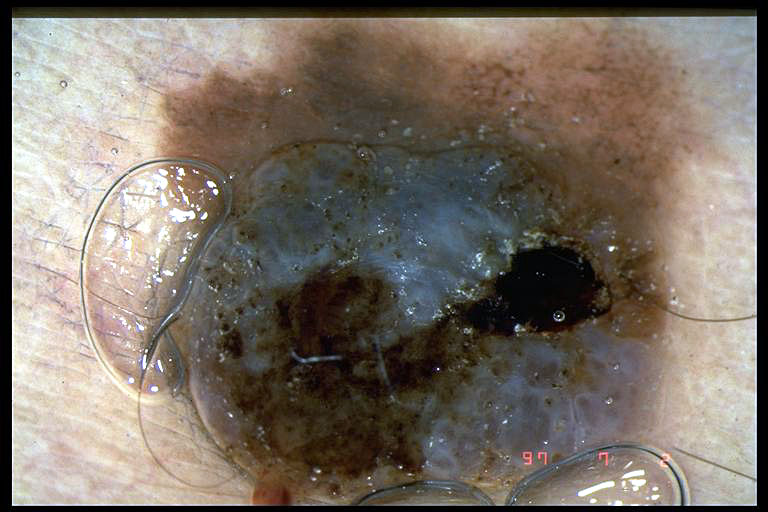}
  \caption{}
  \end{subfigure}

\caption{Artifacts from the out-of-distribution test sets. While (a) PH2 and (b) PAD-UFES-20 present similar artifacts to ISIC2019 (our training set), Derm7pt ((c) Clinical and (d) Dermoscopic) present different ones. We hypothesize this caused debiasing solutions to be more effective in PH2 and~PAD-UFES-20.}
\label{fig:ood-artifacts}
\end{figure}

The differences between artifacts present in training (which are increasingly reinforced as training bias increases) and test may explain such irregular behavior.
Analyzing the artifacts of each out-of-distribution test-set, we verified that the datasets most affected by the debiasing procedures reliably display a subset of the artifacts present on training. 
Specifically, PH2 presents dark corners, while PAD-UFES-20 display ink-markings. Derm7pt present rare cases of dark corners, and different style of rulers. 
Hair is the only artifact in all $4$ test sets, while patches, and gel borders are absent in all sets. In Fig.~\ref{fig:ood-artifacts}, we show a selection of the artifacts from each considered out-of-distribution test-set.
In such scenario, the models appear to learn to avoid known artifacts from training environments instead of learning to rely on clinically-relevant features.

Another possible explanation for such variation is hinted by the overall low performance of NoiseCrop (except for PAD-UFES-20). There is a chance that the low performances are due to the domain shift introduced by the background noise, but this is unlikely since such shift did not affect our ISIC2019 experiments, where NoiseCrop reliably achieved our best performances. 
A more concerning and plausible explanation is that when censored of background information, models can not exploit other available sources of spurious correlations. Such spurious correlations are present in training and may even have very low correlations to the label. In addition, the natural distribution shift of correct features that happen in out-of-distribution sets, cause performances to drop. It is possible then, that the performance achieved by ERM and GroupDRO are overoptimistic.
This shows the challenges of debiasing skin lesion models, agreeing with previous works \cite{bissoto2020debiasing} that suspected models combine weak correlations from several sources that may be hard to detect. For further advancing debiasing, future datasets must explicitly describe possible sources of spurious correlations \cite{daneshjou2022checklist}. 

\noindent\textbf{Summary:} When considering biased training scenarios, our proposed debiasing solutions surpassed baselines in $3$ out of $4$ test sets. Still, improvements depend on the similarity between the confounders used to partition environments and the ones present in test. Models fail when background is censored.

\subsection{Qualitative Analysis}
\begin{figure}[htb]
    \centering
    \begin{subfigure}{0.31\linewidth}
      \includegraphics[width=0.31\linewidth,height=0.3\linewidth]{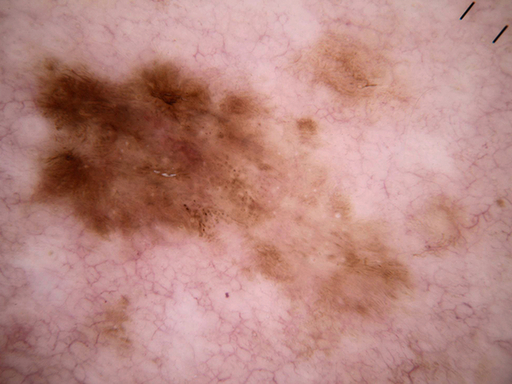}
      \includegraphics[width=0.31\linewidth,height=0.3\linewidth]{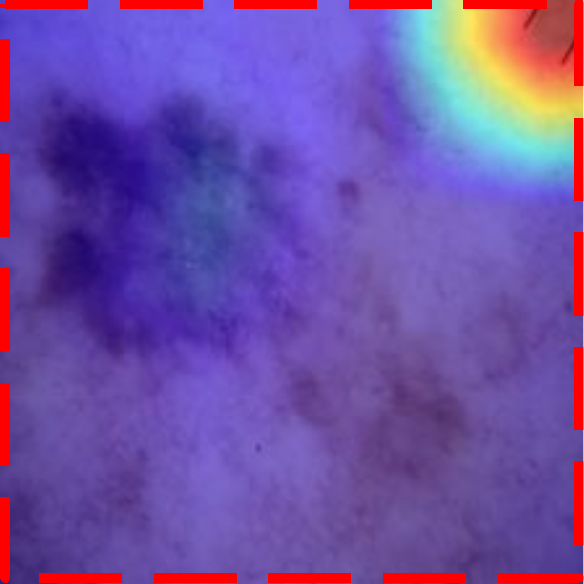}
      \includegraphics[width=0.31\linewidth,height=0.3\linewidth]{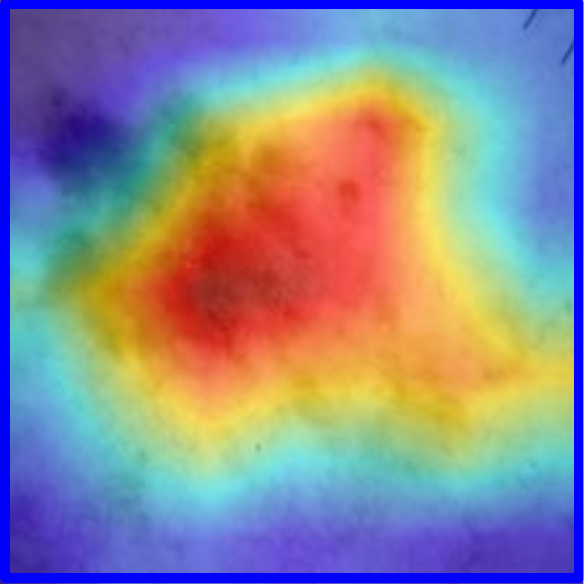}
    \end{subfigure}\hspace{0.15cm}
    \begin{subfigure}{0.31\linewidth}
    \includegraphics[width=0.31\linewidth,height=0.3\linewidth]{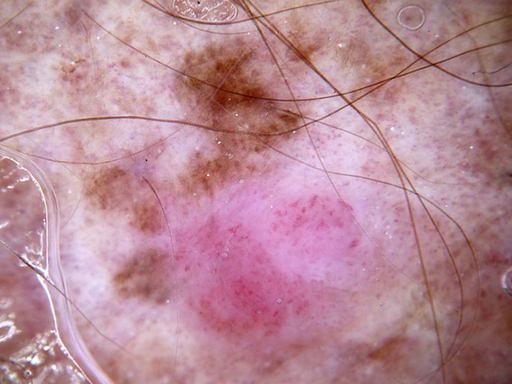} 
      \includegraphics[width=0.31\linewidth,height=0.3\linewidth]{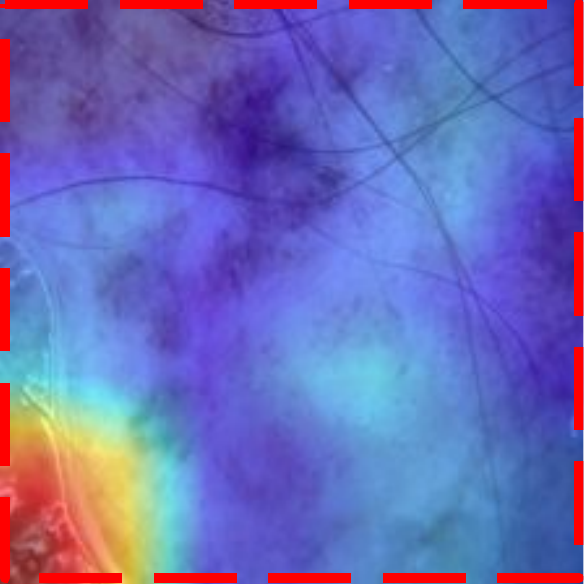}
      \includegraphics[width=0.31\linewidth,height=0.3\linewidth]{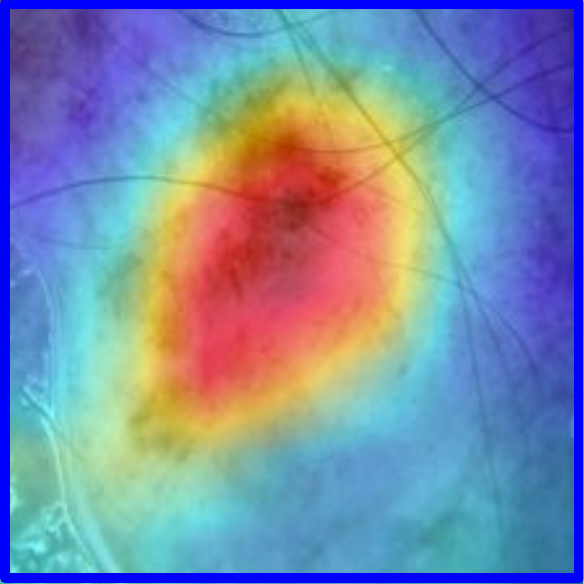}
    \end{subfigure} \hspace{0.15cm}
    \begin{subfigure}{0.31\linewidth}
    \includegraphics[width=0.31\linewidth,height=0.3\linewidth]{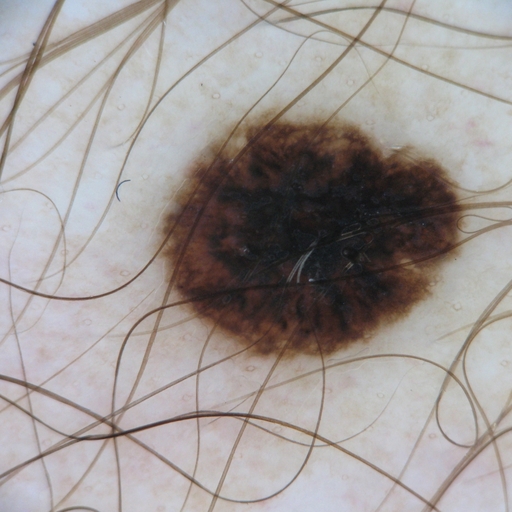}
      \includegraphics[width=0.31\linewidth,height=0.3\linewidth]{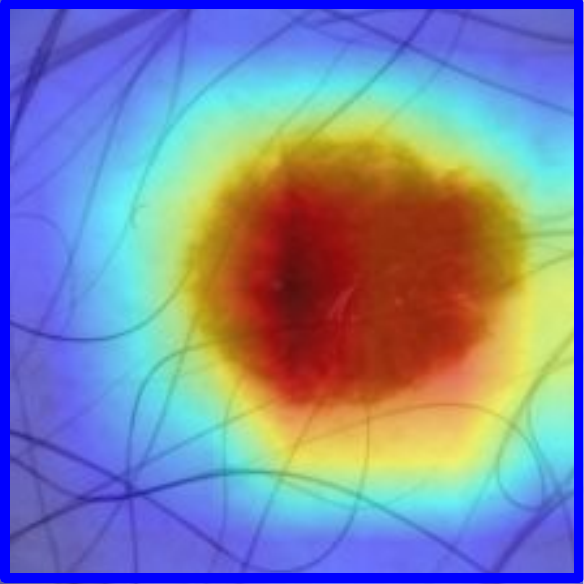}
      \includegraphics[width=0.31\linewidth,height=0.3\linewidth]{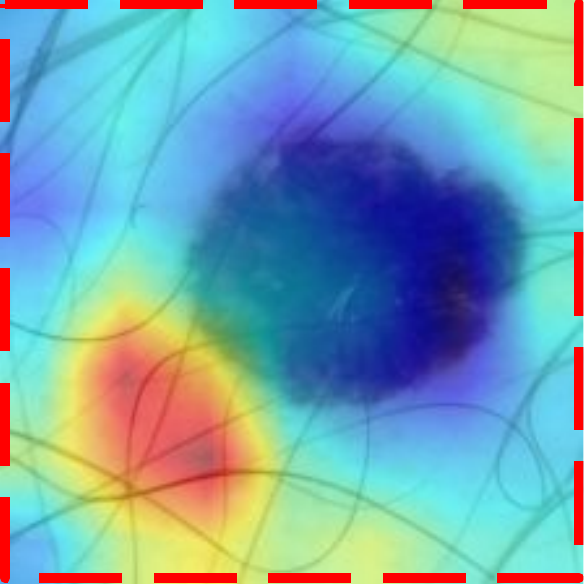}
    \end{subfigure}
    \caption{Qualitative analysis of malignant samples from the trap-test. We show three sets, each showing the original image followed by ScoreCam saliency maps of ERM and GroupDRO models (ours), in this order. Red (dashed) and blue (solid) borders mark wrong and correct predictions, respectively. In most scenarios, GroupDRO can shift the focus of the model from the artifact to the lesion (first two cases). However, there are still failure cases where the opposite happens (last case).}
    \label{fig:scorecam}
\end{figure}

To inspect the effects from another angle, we used ScoreCAM~\cite{wang2020score} to create saliency maps\footnote{To minimize stochastic effects in the saliency maps, we compare models trained with the same random seed.}. We contrast our robust trained model with the ERM solution on the most biased scenario (training bias 1.0). 
In Fig.~\ref{fig:scorecam}, we show cherry-picked malignant cases from the trap-test set that were misclassified by the ERM or GroupDRO, and that focused on an artifact. There are numerous samples in which the saliency maps indicate that ERM models focus on rulers. When trained with GroupDRO, models often correctly shift their attention to the lesion, causing the prediction to be correct. 
There are also cases where the baseline's attention correctly focuses on a lesion (even though the prediction is erroneous) and the robust model focuses on the artifact, but these are considerably less frequent.

\section{Related Work}


\noindent\textbf{Artifacts on skin lesion datasets.}
Artifacts affect skin-lesion-analysis models, which achieve a performance considerably higher than chance in images with the lesion fully occluded \cite{bissoto2019constructing}. Generative models can amplify such biases \cite{mikolajczyk2022biasing}.
Further investigation~\cite{bissoto2020debiasing} analyzed the correlations between artifacts and labels, showing that, even with modest correlations, artifacts harmed performances. An analysis of the ISIC 2019 challenge \cite{combalia2022validation} quantified the error rates of the top-ranked models when artifacts were present, finding that ink markings were particularly harmful for melanoma classification. 
Another work~\cite{daneshjou2022checklist} recommended that future skin-lesion datasets describe artifacts and other potential confounders as metadata.\\

\noindent\textbf{Evaluation of generalization performance.}
Out-of-distribution performance must be measured in challenging protocols, whose craft is laborious requiring attention to class proportions, correlations to other objects, or to background colors, textures, and scenes (e.g.,  ObjectNet \cite{barbu2019objectnet}, ImageNet-A \cite{hendrycks2021natural}). 
Partially or fully synthetic datasets (e.g., natural images on artificial backgrounds) allow fine control of the spurious correlations
and are often employed in more theoretical works, or as a first round of evaluations elsewhere \cite{arjovsky2019invariant, ahmed2020systematic}.
An alternative to synthetic or handcrafted datasets is to employ naturally occurring environments (such as data source) and split the data holding out some environments exclusively for testing \cite{shrestha2021investigation, gulrajani2020search}, e.g., in PatchCamelyon17-WILDS \cite{bandi2018detection}, one of the five source hospitals is used for testing, while the others are used for training. 

Our assessment scheme forgoes either synthetic data or splitting the sets by hand. Our scheme requires (ground-truth or inferred) annotations for potential bias sources (such as the artifacts we use in this work), but once those are available, the trap sets automatically amplify their effect by creating train and test splits with inverse correlations. The tunable trap sets proposed in this work allow controlling a level of bias.\\ 

\noindent\textbf{Debiasing medical imaging.}
Environment-dependent debiasing techniques seldom appear in the literature on medical image analysis. That is partly due to the lack of environment annotations, e.g., potential biasing attributes or artifacts. 
One way to see environments is through the lenses of causality, where they could be thought of as interventions in data \cite{arjovsky2019invariant}. 
Direct interventions in real-world data can be unfeasible or at least uncommon. For example, collecting the same image under different acquisition devices is uncommon if not for domain generalization purpose studies. Other types of shifts, such as the ones characterized by physical attributes, are impossible to intervene upon. It is impossible, for example, to see how a lesion on the face would be if it were in the palms and soles. Still, ideally, we would have enough environments to explain every source of noise in data, with slight differences between them. 
When environments are not annotated \textit{a priori}, works develop mechanisms to create them.
A common strategy is to assign whole data sources as environments \cite{bandi2018detection, koh2020wilds,aubreville2022mitosis}. 
However, when this strategy is successful, the different data sources (and environments) characterize only changes in a few aspects, such as the acquisition device. 
When differences across data sources are considerable, environments differ in many aspects simultaneously, harming debiasing performance.
Other methods to generate environments rely on using differences in classes distributions \cite{yoon2019generalizable}, data augmentation procedures \cite{chang2021stain, zhang2022semi}, or generative modeling \cite{liu2022domain}. 
After environments or domains are artificially generated through one of the techniques above, robust training use environments for feature alignment.

In our work, we use annotations of artifacts to create environments. Each environment presents a unique combination of artifact and label, yielding over $90$ in training environments. Models trained with our environments successfully learned to avoid using artifacts for inference, improving performance in high-bias setups.



\section{Conclusion}
Debiasing skin lesion models is possible. In this paper, we introduced a pipeline that enables bias generalization assessment without access to out-of-distribution sets, followed by a strategy to create environments from available metadata, and finally, a test-set debiasing procedure. We evaluated our pipeline using a large challenging training dataset and noisy (inferred) artifact annotations. 

Our findings suggest that domain generalization techniques, such as GroupDRO, can be employed for debiasing, as long as the environments represent spurious fine-grained differences, such as the presence of artifacts. Also, we showed that models learn a diverse set of features (spurious and robust), even in biased scenarios, and that removing spurious ones during test yields surprisingly good results without any training procedure changes.
When we use training and test-time debiasing, we achieve our best result --- GroupDRO enabled learning more robust features, while NoiseCrop allows using them during inference.
For out-of-distribution sets, the debiasing success depends on the similarity between the artifacts they display and those in training, used to partition environments. Despite potentially learning more robust features with GroupDRO, the presence of different artifacts and spurious correlations in test-time can still bias predictions. 

In future work, we envision methods that are less reliant on labels for both environment partition and test-time debiasing. The domain generalization literature is evolving, proposing methods that learn to separate environments solely from data \cite{creager2021environment, ahmed2020systematic}, but are still to see the same success of supervised approaches.
Alternatively to test-time debiasing, methods for model editing \cite{mitchell2021fast, santurkar2021editing} could enable practitioners to guide models from a few annotated images by making explicit the presence of artifacts and other spurious features. 



\section*{Acknowledgments}
A. Bissoto is funded by FAPESP 2019/19619-7. 
C. Barata is funded by the FCT projects LARSyS (UID/50009/2020) and CEECIND/00326/2017.
E. Valle is partially funded by CNPq 315168/2020-0. 
S. Avila is partially funded by CNPq 315231/2020-3, FAPESP 2013/08293-7, 2020/09838-0, and Google LARA 2021. 
The Recod.ai lab is supported by projects from FAPESP, CNPq, and CAPES.

\clearpage
%
%
\bibliographystyle{splncs04}
\bibliography{egbib}
\end{document}